\documentclass{article}

\usepackage[utf8]{inputenc} % allow utf-8 input
\usepackage[T1]{fontenc}    % use 8-bit T1 fonts
\usepackage{hyperref}       % hyperlinks
\usepackage{url}            % simple URL typesetting
\usepackage{booktabs}       % professional-quality tables
\usepackage{amsfonts}       % blackboard math symbols
\usepackage{nicefrac}       % compact symbols for 1/2, etc.
\usepackage{microtype}      % microtypography
\usepackage{xcolor}         % colors
\usepackage{amsmath,amsthm}
\usepackage{amsfonts}
\usepackage{amssymb}
\usepackage{mathtools}
\usepackage{algorithm}
\usepackage{color}
\usepackage{subcaption}
\usepackage{wrapfig}
\usepackage{authblk}

\usepackage{fullpage}
\usepackage[utf8]{inputenc}
\usepackage[T1]{fontenc} 
\usepackage{booktabs}
\usepackage{amsmath,amsthm}
\usepackage{amsfonts}
\usepackage{amssymb}
\usepackage{mathtools}
\usepackage{algorithm}
\usepackage{algpseudocode}
\usepackage{color}
\usepackage{subcaption}
\usepackage{hyperref}
\usepackage{wrapfig}
\usepackage[numbers]{natbib}
\usepackage{authblk}

\usepackage{enumitem}
\setlist[itemize]{topsep=0pt,itemsep=-0.4ex,partopsep=0.8ex,parsep=0.8ex}

  \usepackage{color-edits}

\usepackage{algpseudocode}
\usepackage{soul}

\newcommand{\xhdr}[1]{\vspace{2mm} \noindent{\bf #1}}

\newcommand{\dss}{\mathbf{Z}}

\newtheorem{thm}{Theorem}[section]

\theoremstyle{definition}
    \newtheorem{defn}[thm]{Definition}

\author[1]{Shuai Tang\footnote{
Shuai and Steven are the lead authors, and other authors are ordered alphabetically. shuat@amazon.com}}
\author[1,3]{Zhiwei Steven Wu$^\ast$}

\author[1]{\\ Sergul Aydore}
\author[1,2]{Michael Kearns}
\author[1,2]{Aaron Roth}

\affil[1]{Amazon AWS AI/ML}
\affil[2]{University of Pennsylvania}
\affil[3]{Carnegie Mellon University}
\date{}

\title{Membership Inference Attacks on Diffusion Models \\via Quantile Regression}

\begin{document}
\maketitle

\begin{abstract}
Recently, diffusion models have become popular tools for image synthesis because of their high-quality outputs. However, like other large-scale models, they may leak private information about their training data. Here, we demonstrate a privacy vulnerability of diffusion models through a \emph{membership inference (MI) attack}, which aims to identify whether a target example belongs to the training set when given the trained diffusion model. Our proposed MI attack learns quantile regression models that predict (a quantile of) the distribution of reconstruction loss on examples not used in training. This allows us to define a granular hypothesis test for determining the membership of a point in the training set, based on thresholding the reconstruction loss of that point using a custom threshold tailored to the example. We also provide a simple bootstrap technique that takes a majority membership prediction over ``a bag of weak attackers'' which improves the accuracy over individual quantile regression models. We show that our attack outperforms the prior state-of-the-art attack while being substantially less computationally expensive --- prior attacks required training multiple ``shadow models'' with the same architecture as the model under attack, whereas our attack requires training only much smaller models. 

\end{abstract}

\section{Introduction}

Diffusion models, based on generative neural networks, have gained attention in the field of image generation \cite{ho2020denoising, song2019generative}. It has been shown that diffusion models are remarkably capable of generating images that are higher-quality than previous approaches such as GANs and VAEs, while also being more scalable. However, as the size of these models has grown drastically over the last decade, so has the privacy concern that these large-scale diffusion models may reveal sensitive information about the dataset they are trained on.

One of the most popular classes of methods to evaluate the privacy risks of machine learning (ML) models is \emph{membership inference (MI) attacks} (e.g., \cite{Homer+08, ShokriSSS16, Jayaraman2019, JagielskiUO20, NasrSTPC21, CarliniCNSTT22}), in which an attacker aims to determine if a target example belongs to the training dataset given the trained model. 
The success of a MI  attack falsifies privacy protections on the existence of any individual, i.e. differential privacy guarantees \cite{DworkR14}. 
MI can also be disclosive if e.g. membership in the training data is determined based on a sensitive attribute (as it would if e.g. the dataset consisted of medical records for patients with a particular disease). 
  In addition, MI attacks can be a building block for other more sophisticated attacks such as \emph{extraction attacks} on generative models. 
In general, a successful MI attack with reasonable side information is a strong indicator of a privacy vulnerability. Finally, when applied to differentially private algorithms \cite{DworkMNS06}, MI attacks can serve as privacy auditing tools by providing lower bounds on the privacy parameters, which in turn assess the tightness of the privacy analyses \cite{JagielskiUO20, NasrHSBTJCT23} and help identify potential errors in the privacy proof or implementation \cite{DP-Debug, StadlerOT22}.

A majority of the existing MI attacks focus on supervised learning \cite{YeomGFJ18, WangJEG19,JayaramanWEG20,NasrSTPC21, ShokriSSS16, CarliniCNSTT22}, and there has been significantly less development on  MI attacks against generative models (e.g., \cite{LOGAN, BreugelSQS23,carlini2023extracting}). The goal of our work is to develop strong MI attacks against state-of-the-art diffusion models.

Our work extends the quantile-regression-based attacks in \cite{QMIA} for supervised learning to attacks for diffusion models. For a given trained diffusion model parameterized by $\theta$, our attack first learns a quantile regression model on public auxiliary data that predicts the $\alpha$-quantile $q_\alpha(z)$ of  $\theta$'s reconstruction loss on each example $z$ (formally defined in Definition~\ref{t-error}). Then we indicate an example is a member of the training set if its reconstruction loss is lower than its predicted $\alpha$-quantile. By design, the attack has a false positive rate of $\alpha$: that is the probability that it incorrectly declares a randomly
selected point $z$ that was not used in training to have been used in training is $\alpha$. We further boost the attack performance of our approach through bagging aggregation over small quantile regression models. 
We evaluate our attack on diffusion models trained on image datasets, and demonstrate four major advantages:
\begin{itemize}
    \item Our quantile-regression-based attack obtains state-of-the-art accuracy on several popular vision datasets. Even though our attacks leverage the same reconstruction loss function considered in \cite{pmlr-v202-duan23b}, their attack leverages the same \emph{marginal approach} in \cite{YeomGFJ18} that applies a 
 uniform threshold (that is, the $\alpha$-quantile on the marginal distribution over the reconstruction loss) across all examples. In comparison, our attack is \emph{conditional} since it applies a finer-grained per-example threshold when performing membership inference. 

    \item Compared to the prior state-of-the-art MI attacks against diffusion models \cite{pang2023white} also in the white-box setting, we achieve higher accuracy without suffering their computational cost. Similar to the Likelihood Ratio Attack (LiRA) attack proposed by \cite{CarliniCNSTT22}, the Gradient-Subsample-Aggregate (GSA) attack 
    in \cite{pang2023white} requires training multiple \emph{shadow models}, each of which is obtained by running the training algorithm of the diffusion model under attack on a randomly drawn dataset. While the accuracy of the MI attack improves as the number of shadow models increases, their approach also becomes computationally prohibitive. In comparison, our approach only requires learning \emph{tiny} quantile regression models.

    \item Since our attack does not rely on shadow models, 
    it also requires significantly fewer details about the training algorithm of the diffusion model under attack, such as hyperparameters and network architecture used in training. In fact, our attack is effective even though the neural network for the quantile regression model has significantly fewer parameters than the attacked diffusion models.

    \item While both our work and the prior work of \cite{QMIA} rely on quantile regression for MI attacks, an important distinction in ours is the use of bootstrap aggregation --- \emph{bagging} --- that takes an ensemble of tiny quantile regression models, namely a ``bag of weak attackers.'' Bagging generally improves the attack performance by reducing the variance of the individual models, each of which can be viewed as a weak hypothesis test. The use of bagging immediately enriches the space of MI attacks.   
    Specifically, we use bagging techniques in conjunction with small quantile regression models, which leads to a substantial improvement in accuracy by introducing little computational overhead (due to the small model size). In comparison, the MI attack in \cite{QMIA} primarily leverages a parametric approach to fit a single quantile regression model, and it falls short in performance compared to the shadow models approach \cite{CarliniCNSTT22} on datasets such as CIFAR10.

\end{itemize}

\section{Background and Preliminaries}
We here present the objective of a Membership Inference (MI) attack along with required side information, and briefly introduce diffusion models for  context.

\subsection{Membership inference attacks}\label{sec:miabackground}

Membership inference (MI) is a common privacy attack that attempts to predict whether a given example was used to train a machine learning model (e.g., ~\cite{Homer+08, YeomGFJ18, ShokriSSS16, JagielskiUO20,JayaramanWEG20,NasrSTPC21,CarliniCNSTT22}). Our work focuses on performing MI attacks on diffusion models.

\textbf{Problem statement.} Given a training dataset $\dss$ 
drawn from an underlying distribution $P$, a diffusion model $\theta$ is trained on $\dss$. The goal of a membership inference attack is to infer whether a target example $z^*$ 
was included in the training set $\dss$ or not. 
  
\textbf{Adversary's side information.}  Similar to almost all prior work on membership inference \cite{CarliniCNSTT22, pmlr-v202-duan23b, QMIA, ShokriSSS16}, we assume the adversary has access to some public data drawn from $P$. In the standard terminology of MI, there are two types of access to the algorithm's output. In a \emph{black-box} attack, the adversary only has access to the generated synthetic data. In a \emph{white-box attack}, the adversary has access to the generative model $G$. In this work, we focus on white-box attacks, and specifically we only need access to the parameters of the trained diffusion model, without information regarding the training algorithm.

\subsection{Diffusion Models}
Before describing our attack, it is helpful to briefly describe how diffusion models work at a high level, following the notation of \cite{ho2020denoising}.
For a real image, a diffusion model provides a stochastic path from the image to noise. A diffusion model consists of two processes: 
\begin{enumerate}
    \item a $T$-step diffusion process (denoted as $q$ below) that iteratively adds Gaussian noise to an image, and
    \item a denoising process (denoted as $p_\theta$ below) that gradually reconstructs the image from noise. 
\end{enumerate}
Let $z_0$ be the real image without noise and $z_T$ be the noisy image with the largest amount of noise. The transitions of diffusion and denoising are mathematically described as:
\begin{align}
q({z_t|z_{t-1}})=\mathcal{N}(z_t; \sqrt{1-\beta_t}z_{t-1},\beta_t I) \\
p_\theta(z_{t-1}|z_t)=\mathcal{N}(z_{t-1}; \mu_\theta(z_t,t), \Sigma_\theta(z_t,t))
\end{align}
where $q_{z_t|z_{t-1}}$ is the probability distribution of the diffused image $z_t$ given the previous image $z_{t-1}$, $p_\theta(z_{t-1}|z_t)$ is the probability distribution of the denoised image $z_{t-1}$ given the noisy image $z_t$, $\mu_\theta(\cdot)$ and $\Sigma_\theta(\cdot)$ are the mean and covariance of the denoised image, respectively, as parameterized by the model parameters $\theta$ 
, and $\beta_t$ is a noise schedule that controls the amount of noise added at each step. Moreover, the marginal distribution at any time step $t$ given the example $z_0$ can be written as 
\begin{align}
    q(z_t \mid z_0) = \mathcal{N}(z_t; \sqrt{\overline{\alpha}_t} z_0, (1- \overline{\alpha}_t)I),
\end{align}
where $\alpha_t = 1 - \beta_t$ and $\overline{\alpha}_t = \prod_{s=1}^t \alpha_s$. We will work with the following re-parameterization of $\mu_\theta$ with 
\begin{align}
\mu_\theta(z_t, t) = \frac{1}{\sqrt{\alpha_t}} \left( z_t  - \frac{\beta_t}{1 - \overline{\alpha}_t} \epsilon_\theta(z_t, t) \right)
\end{align}
where $\epsilon_\theta$ is a predictor (given by $\theta$) that predicts the noise component given $z_t$.

\xhdr{Loss function.} Many MI attacks proceed by identifying a loss function and making membership inference by comparing the loss on the target example with a threshold. Intuitively, if the loss is unusually low, then there is evidence that the example was part of the training set. For supervised learning models, MI attacks typically leverage the classification loss (e.g., the cross-entropy loss). For diffusion models, existing work has proposed candidates of loss functions that measure the reconstruction error at different time steps of the diffusion process \cite{carlini_extracting_2020, pmlr-v202-duan23b}. We leveraged the $t$-error function defined in \cite{pmlr-v202-duan23b}, which has the compelling advantage that it is deterministic and avoids repeated sampling from the diffusion process. Consider the following deterministic approximation of the diffusion and denoising processes:
\begin{align}
    z_{t+1} &= \phi_\theta(z_t, t ) = \sqrt{\overline{\alpha}_{t+1}} f_\theta(z_t ,t) + \sqrt{1 - \overline{\alpha}_{t+1}} \epsilon_\theta(z_t, t)\\
    z_{t-1} &= \psi_\theta(z_t, t) = \sqrt{\overline{\alpha}_{t-1}} f_\theta(z_t, t) + \sqrt{1 - \overline{\alpha}_{t-1}} \epsilon_\theta(z_t, t)
\end{align}
where $f_\theta(z_t, t) = \frac{z_t - \sqrt{1- \overline{\alpha}_t} \epsilon_\theta(z_t, t)}{\sqrt{\overline{\alpha}_t}}$ is the estimate of $z_0$ given the $z_t$ and the prediction $\epsilon_\theta(z_t, t)$. Then we could also define the deterministic 
reverse result as
\begin{align}
\Phi(z_0, t) = \phi_\theta(\cdots \phi_\theta(\phi_\theta(z_0, 0), 1) \ldots , t-1)
\end{align}
Now we can define the reconstruction loss function, termed as $t$-error, that is used in our MI attack.  
\begin{defn} ($t$-error)\label{t-error}
For a given sample $z_0$ and the deterministic reverse result $\tilde{z}_t=\Phi_\theta(z_0, t)$ 
at time step $t$, the approximated posterior estimation error at step $t$ is defined as $t$-error:
\begin{align}
\hat{\ell}_t({\theta,z_0})=||\psi_\theta(\phi_\theta(\tilde{z}_t, t),t) - \tilde{z}_t ||^2,
\end{align}
\end{defn}
where $\hat{\ell}_t$ is estimated from the deterministic reverse function that is used to approximate the sampling process in the Markov Chain, and $\tilde z_t$ refers to the deterministic outcome. 
Intuitively, the $t$-error function measures how much we change $\tilde{z}_t$ if we take one step in the deterministic diffusion process $\phi_\theta$ and then rewind back with one step of deterministic  denoising $\psi_\theta$. While this loss function is not what the training algorithm optimizes, it provides a deterministic approximation to the loss function during training \cite{pmlr-v202-duan23b, ho2020denoising}. Thus, smaller $t$-error values provides evidence that $z_0$ was used to train the diffusion model $\theta$.

\section{MI Attacks with Quantile Regression}

We will now describe our new membership inference attacks. Under the setting in Sec.~\ref{sec:miabackground}, we assume that the attacker has access to a set of public examples $D$ drawn from the underlying distribution $P$. Given the public dataset $D$, a choice of $t$ for the $t$-error function, and the trained diffusion model $\theta$, the attacker learns a quantile regressor $q_\alpha$ such that $q_\alpha(z)$ 
predicts the $\alpha$-quantile of the $t$-error $\hat \ell_t(\theta, z)$ for each example $z$ in $D$, where $\alpha$ is a parameter that controls the false-positive rate. Then on any target example $z^*$, the attacker declares the example is a member of the training set if and only if the $t$-error $\hat \ell_t(\theta, z^*) \leq q_\alpha(z^*)$. The formal description of the algorithm is in Algorithm \ref{alg:qmia-synth}.

By design, our attack has a {false-positive rate} of $\alpha$, which is the probability that an attacker incorrectly declares a randomly selected point $z$ that was not used in training to have been used in training is $\alpha$. By varying the parameter $\alpha$, we can then trace the trade-off curves of \emph{true-positive rates} at different false-positive rates.

\begin{algorithm}[t]
\caption{Quantile Regression MI attacks for Diffusion Model }
\label{alg:qmia-synth}
\begin{algorithmic}
\Require A set of auxiliary examples $D$ drawn from $P$, target example $z^*$, trained diffusion model $\theta$, a choice of $t$ for $t$-error function. Target false-positive rate $\alpha$.
\For{each $z\in D$}
    \State evaluate the score $\hat \ell_t(\theta, z)$
\EndFor
\State 
Learn a quantile regression model $q_\alpha$ such that $q_\alpha(z)$ predicts the $\alpha$-quantile of the score $\hat\ell_t(\theta, z)$ conditioned on $z$. \\
\Return "IN" if $\hat \ell_t(\theta, z^*)\leq q_\alpha(z^*)$, otherwise "NO"
\end{algorithmic}
\end{algorithm}
\begin{algorithm}[t]
\caption{Bag of Weak Attackers}
\label{alg:bag-of-weak-attackers}
\begin{algorithmic}
\Require A set of auxiliary examples $D$ drawn from $P$, target example $z^*$, trained diffusion model $\theta$, a choice of $t$ for $t$-error function, target false-positive rate $\alpha$, and the number of weak attackers $m$.
\State Initialize vote=$0$
\For{each $z\in D$}
    \State evaluate the score $\hat \ell_t(\theta, z)$
\EndFor
\For{$i \in [m]$}
    \State Bootstrap sampling a dataset $D_i$ from $D$ (Sample with replacement, and $|D_i|=|D|$)
    \State Learn a $\alpha$-quantile regression model $q_{\alpha,i}$ on $D_i$ (such that $q_{\alpha,i}(z)$ predicts the $\alpha$-quantile of the score $\hat\ell_t(\theta, z)$ conditioned on $z$.)
    \State $\text{vote}=\text{vote}+1$ if $\hat \ell_t(\theta, z^*)\leq q_{\alpha,i}(z^*)$
\EndFor \\
\Return "IN" if $\text{vote} \geq \frac{m}{2}$, otherwise "NO"
\end{algorithmic}
\end{algorithm}

\subsection{Quantile Regression Learner}
A generic way to train a quantile regression model is to optimize pinball loss over some function class $\mathcal{Q}$ (e.g., neural networks). Formally, for any observed $t$-error $\hat \ell$ and quantile prediction from $q_\alpha$ at a target level $\alpha$, the pinball loss is defined as
\begin{align}
L_{\alpha}(\hat \ell, q_\alpha) = (q_\alpha - \hat \ell)(\mathbf{1}[\hat\ell \leq q_\alpha] - \alpha)
\end{align}
where $\mathbf{1}(\cdot)$ is an indicator function 
Then we can find a quantile regression model $q_\alpha(\cdot)$ that  minimizes the \emph{pinball loss}:
\begin{align}
\min_{q_\alpha\in \mathcal{Q}} \sum_{z\in D} L_{\alpha}(\hat\ell_t(\theta, z), q_\alpha(z)), 
\end{align}
where $\mathcal{Q}$ is the class of quantile regression models. The pinball loss is minimized by the function that predicts for each $z$ the target $\alpha$-quantile of the $t$-error conditioned on $z$. However, prior work \cite{QMIA}  has found that pinball loss tends to be a difficult loss function to minimize. 

\begin{figure*}[t]
    \centering
    \includegraphics[width=\textwidth]{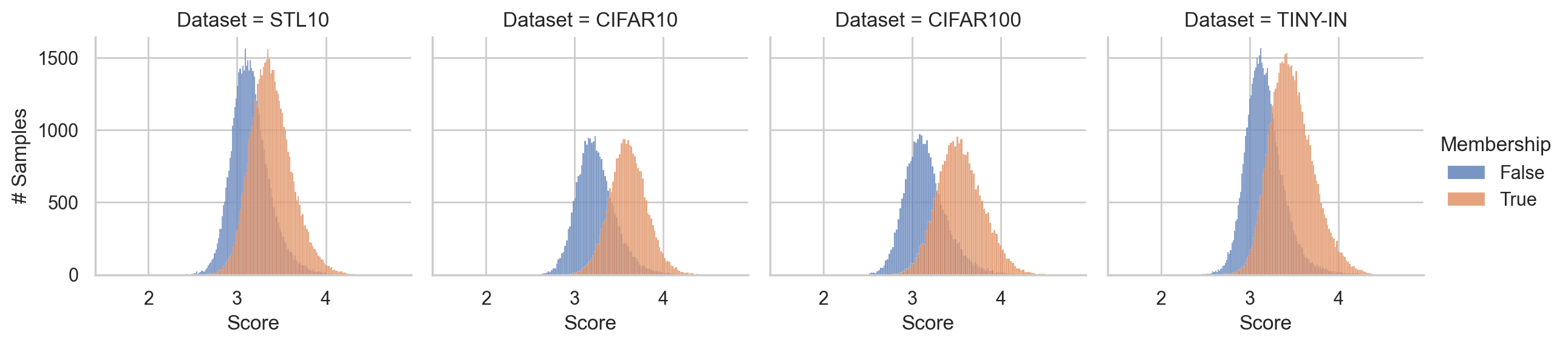}
    \caption{Distrbutions of the (negative) log transformation of the $t$-error on members and nonmembers of a dataset. It is clear that on each dataset, members and nonmembers have slightly different marginal score distributions, however, they are not drastically different from each other, which explains why the marginal baselines are not optimal, and also motivates our approach that conditions the score prediction on the input sample. }
    \label{fig:score_dist}
\end{figure*}

\subsection{Bag of Weak Attackers}

One of the challenges, as is also mentioned in \cite{QMIA}, is that sometimes directly optimizing pinball loss is difficult in practice since the gradient is either $-\alpha$ or $1-\alpha$ depending on whether the quantile prediction is smaller or larger than the target. Rather than tackling the optimization challenge directly, 
we take advantage of the fact that the quantile regression model here can be a much smaller neural network than the targeted diffusion models, which enables us to train multiple models with minimal computational overhead. In particular, we adopt the bagging approach \cite{breiman1996bagging} to improve the generation of our attack by taking an ensemble of tiny models trained on bootstrapped datasets. Since individual weak models may have relatively poor attack performance,
and bagging improves the performance, we call our approach ``bag of weak attackers''. 
After learning, each weak attacker makes a binary decision on an input sample to decide whether this sample is used in training the diffusion model or not, and we simply take the majority vote over all weak attackers to obtain the final decision. The formal description of the algorithm is in Alg. \ref{alg:bag-of-weak-attackers}. It is worth noting that quantile regression models only need to be trained once before attacking any sample, and can be used to attack many samples since its computational cost is amortized over all samples under attack.

In Section~\ref{sec:exp}, we empirically demonstrate the performance improvement from bagging, and the success of our attack using a bag of weak attackers, which are efficient tiny models that are fast to learn. 

\section{Experimental Details}
\label{sec:exp}
We demonstrate the effectiveness of our membership inference attack via quantile regression on four de-noising diffusion probabilistic models \cite{ho2020denoising} (DDPMs) trained on CIFAR-10, CIFAR-100 \cite{krizhevsky2009learning}, STL-10 \cite{coates2011analysis} and Tiny-ImageNet, respectively. On each dataset, data samples are split into two halves, and one half is regarded as the private samples $\dss$ for training a DDPM. The other half is then split into two sets, including one as the public samples $D$ that are auxiliary information, and the other as the holdout set for testing. On public samples, we train quantile regression models. 

The base for our quantile regression model is a ResNet model, and it is attached with multiple prediction heads, each of which predicts target $\alpha$-quantile for a specific value of $\alpha$. Compared to the standard ResNet-18 model for classifying CIFAR-10, we design each attacker to be much smaller than the standard ResNet model. We present results with a varying number of weak attackers, each with a varying number of parameters.

In our experiments, we use the same diffusion models trained on CIFAR10 and CIFAR100 released by \cite{pmlr-v202-duan23b}, and trained our own diffusion models --- specifically DDPM --- on STL10 and Tiny-ImageNet using lower resolution, $32 \times 32$, due to the limit of computing resources. Each diffusion model was trained with 80k steps, and it took around 2 days to finish training on a single V100 GPU card. After obtaining these trained diffusion models, for membership inference attacks, we use a fixed $t = 50$ in the $t$-error function. \cite{pmlr-v202-duan23b} suggested that the choice of $t$ does not influence the results drastically.

We adopt the same evaluation metric as prior work \cite{QMIA,CarliniCNSTT22}. Specifically, we are interested in the True Positive Rates (TPRs) at very low False Positive Rates (FPRs). Intuitively, a successful membership inference attack should identify true members with high accuracy, and in the meantime, make few mistakes on declaring nonmembers as members.

\subsection{Comparison Partners}
We mainly compare our MI attacks via quantile regression with two approaches. The first one is a simple \textbf{marginal} baseline, which only looks at the error distributions of members and nonmembers, respectively, and for a target FPR value $\alpha$, it computes the quantile on $t$-errors of the public samples, and then the performance of this marginal baseline is evaluated on the private samples and the holdout set. It is clear that the marginal baseline only produces a single threshold for a target FPR, and it does not condition on the input images, whereas ours learns to predict the threshold for a given image, thus each images has a different threshold for a target FPR. Fig. \ref{fig:score_dist} illustrates the error distributions on individual datasets, and it is clear that the error distribution of members is different from that of nonmembers, but using a uniform (single) threshold for all samples would lead to poor performance. 
Our attack produces, for a fixed $\alpha$-value, a sample-conditioned threshold by learning quantile regression models, thus, for each sample under attack, the threshold is unique, which empowers the membership inference attack.

The other comparison partner is also a white-box attack using LiRA with gradient information \cite{pang2023white}, namely \textbf{GSA}. 
The \textbf{LiRA} attack formulates MI as hypothesis testing. Let $\theta$ denote the trained generative model, then the two competing hypotheses are 
\begin{align}
H_0: \theta\sim A(\dss) \mid  z^* \not\in \dss 
\qquad\mbox{ and }\qquad H_1: \theta \sim A(\dss)\mid  z^* \in \dss. 
\end{align}
These hypotheses correspond to whether or not the input dataset $\dss$ includes the target example $z^*$. 
Despite the simple formulation, estimating the two distributions requires training shadow models using random subsets from the same data domain, and each shadow model is a diffusion model, which may take days to train. 

Instead of looking at the distribution over the algorithm's output,
we instead consider the distribution over examples with and without conditioning on the fixed observed output $\theta$ returned by a training algorithm. Compared with unconditioned hypothesis testing approaches, including LiRA, this construction forces the test to be model-dependent. 
Formally, given the target point $z^*$ and the trained generative model $\theta$, there are two hypotheses:
\begin{align}\label{NP-2}
H_0 \colon z^* \sim P \mid A(\dss) = \theta 
\qquad \mbox{ and } \qquad H_1 \colon z^* \sim \dss \mid A(\dss) = \theta
\end{align}
That is, $H_0$ asserts 
that $z^*$ is a random example drawn from the underlying distribution $P$, but $H_1$ asserts $z^*$ is a random example drawn from a dataset $\dss$ that produces the trained model $\theta$. 
In this view, MI is about distinguishing two distributions over examples instead of over trained models. Distinguishing distributions over trained models, e.g. LiRA, requires learning numerous shadow models using subsets of public data from the same domain. By conditioning the test on the target model in the above setup, MI attack then only needs to train a quantile regression model that predicts the quantile of the model score distribution for each example, rather than numerous shadow models. If the model score of the target example $z^*$ is lower than its predicted $\alpha$-quantile, then the attack indicates the example was used in the training set. By design, the attack aims to achieve achieve a false-positive rate of $\alpha$ (e.g., 5\%).

The advantages of our attack is that, firstly, our hypothesis testing setup takes the condition on the target model, which makes our attack model-specific; secondly, our attack only requires learning a bag of tiny models, which is much more computationally efficient. 

Regarding the quantile-regression-based attack in \cite{QMIA}, which has a similar concept as ours but for attacking classification models, the main difference is that ours doesn't require using the surrogate Gaussian likelihood objective or hyperparameter optimization (HPO), which doesn't have a fixed training time for an attack,
and ours only requires bagging over tiny neural networks without HPO. We speculate that the bagging method over tiny networks works well in our setting because of the choice of reconstruction error ($t$-error in Definition \ref{t-error}).

\begin{table*}[t]
\caption{Performance of MI Attacks on CIFAR10 and CIFAR100. Each weak attack is a neural network with only 5,666 parameters, and the number of weak attackers is 7.}
\label{tab:cifar10_cifar100}
\resizebox{\textwidth}{!}{
\begin{tabular}{c|cc|cc}
\toprule
    Dataset     & \multicolumn{2}{c}{CIFAR-10}    & \multicolumn{2}{c}{CIFAR-100}   \\
\midrule        
   MI Attack       & TPR @ 1\% FPR & TPR @ 0.1\% FPR & TPR @ 1\% FPR & TPR @ 0.1\% FPR \\
\hline        
Bag of Weak Attackers & 99.94\%       & 99.86\%         & 99.89\%       & 99.75\%         \\
\hline
GSA$_1$ (Shadow Models) & 99.70\%       & 82.90\%         & -       & -         \\
GSA$_2$ (Shadow Models) & 97.88\%       & 58.57\%         & -       & -         \\
\hline
Marginal Baseline & 9.6\%         & 0.7\%           & 11.06\%       & 5.76\%         \\
\bottomrule
\end{tabular}
}
\end{table*}
\begin{table*}[t]
\caption{Performance of MI Attacks on Tiny-ImageNet and STL-10}
\label{tab:tinyin_stl10}
\resizebox{\textwidth}{!}{
\begin{tabular}{c|cc|cc}
\toprule
   Dataset      & \multicolumn{2}{c}{Tiny-ImageNet}     & \multicolumn{2}{c}{STL-10}      \\
\midrule         
   MI Attack      & TPR @ 1\% FPR & TPR @ 0.1\% FPR & TPR @ 1\% FPR & TPR @ 0.1\% FPR \\
\hline
Bag of Weak Attackers & 99.99\%        & 99.98\%         & 99.98\%       & 99.98\%         \\
\hline
Marginal Baseline & 8\%           & 0.32\%          & 5.78\%        & 0.55\%         \\
\bottomrule
\end{tabular}}
\end{table*}
\subsection{Main Results}
Numerical results are presented in Table \ref{tab:cifar10_cifar100} and Table \ref{tab:tinyin_stl10}, and they are averaged across 10 random seeds. The weak attacker in our experiment is a small neural network with only 5,666 parameters, and the number of weak attackers is 7.
We can see that our attack on CIFAR10, besides being much more efficient, outperforms GSA attacks when we focus on lower TPR (0.1\%). We also have demonstrated the effectiveness of our attack on diffusion models trained on other image datasets, including STL10 and Tiny-ImageNet in Table \ref{tab:cifar10_cifar100} and \ref{tab:tinyin_stl10}. Besides the performance improvement over prior work, our algorithm is also computationally efficient and requires no knowledge about the training algorithm of the diffusion models.

\subsection{Time Consumed in Preparing An Attack}
Apart from simply using a uniform threshhold on the marginal distribution, other white-box attack approaches including ours, involve learning machine learning models. Efficient attack algorithms enable attackers to launch attack more frequently on a machine learning system, and, on the bright side, it also enables frequent privacy auditing for people who are maintaining these systems. 
\begin{table*}[t]
\caption{Time Consumed in Preparing An Attack on Tiny-ImageNet.}
\label{tab:clocktime}
\resizebox{\textwidth}{!}{
\begin{tabular}{l|l|l}
\toprule
Preparation Steps                    & \begin{tabular}[c]{@{}l@{}}Computing Scores on Public Samples\\ using the Target Model\end{tabular} & Learning Models                                                                      \\
\midrule
GSA (6 Shadow models)                & 0 mins ( not required )                                                                             & \begin{tabular}[c]{@{}l@{}}2 days (x 6 shadow models = 12 days)\end{tabular}      \\
Quantile (8 regression models ) & 8 mins                                                                                              & \begin{tabular}[c]{@{}l@{}}4 mins (x 8 regression models = 32 mins )\end{tabular} \\
\bottomrule
\end{tabular}
}
\end{table*}
LiRA attack on a target diffusion model trains shadow models, of which each is a diffusion model of the same size. \cite{pang2023white} proposed to extract gradient information, and it introduces additional latency overhead on top of learning shadow models. On contrary, our approach learns simple quantile regression models. To demonstrate the efficiency of our algorithm in preparing the attack, we estimated the clock time on a single V100 GPU card, and the rough estimates are presented in Table \ref{tab:clocktime}. 

\subsection{Impact of Bagging}
Since our algorithm doesn't require knowledge of the training algorithm of the target diffusion model, the quantile regression model technically can be an arbitrary machine learning model. Therefore, with an inappropriately-chosen model architecture for quantile regression, the attack performance at low FPR may not be desirable. Our proposal to alleviate this issue to through bagging, hence namely ``bag of weak attackers''. We now present empirical ablation study of the impact of bagging on various model sizes. 

We select the base architecture for our quantile regression as a ResNet model, however, we significantly reduce the number of channels in each layer to make it much smaller in terms of the number of parameters and make it much more efficient to train and conduct inference. A standard ResNet model with 18 layers for the classification task on CIFAR10 has around $4.4\times 10^7$ parameters, and in our experiments, the smallest ResNet for quantile regression has only $5.6\times 10^3$ parameters. Our way of reducing the number of parameters is through reducing the number of channels in each convolution layer by a specific factor. 

We consider three configuration of the number of channels in ResNet, which then results in three base models with different number of parameters, including $5.6\times 10^3$, $2.0\times 10^4$, and $8.0\times 10^4$. It is worth noting that the number of parameters in a single diffusion model in our experiments is $7.1 \times 10^8$, and our attack models are significantly smaller than the target model. We then vary the number of weak attackers from 1 to 7. Each configuration is run with 10 random seeds.

\begin{figure*}[t]
    \centering
    \includegraphics[width=\linewidth]{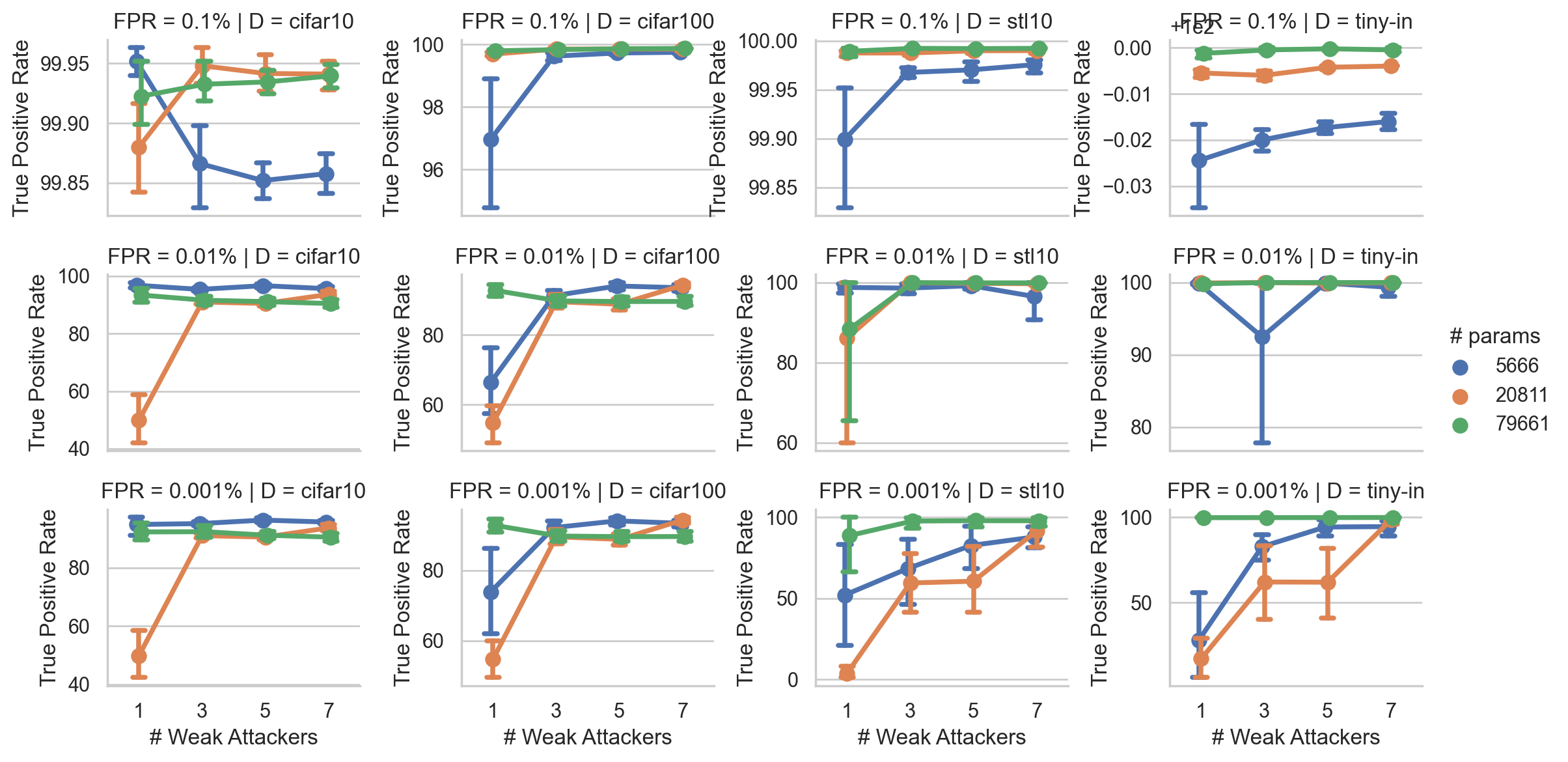}
    \caption{The Effect of Bagging. The bagging predictor takes the majority vote over the decisions made by weak attackers, which in our case, are quantile regressors. Due to the variance reduction aspect of bagging, we expect the attack performance inceases as the number of weak attackers increases. We present the effect of bagging at three different fixed FPR values, including 0.1\%, 0.01\%, and 0.001\%, on four diffusion models, and we also present the impact on three neural networks of different sizes. Overall, we can see that the attack performance demonstrates a non-decreasing trend as the number of weak attacker increases. }
    \label{fig:bagging}
\end{figure*}

\paragraph{Improving Attack Performance. } Figure \ref{fig:bagging} presents results that indicate the effectiveness of our ``bag of weak attackers''. We can see that, for the smallest model, essentially the weakest attacker, increasing the number of attackers improves the performance drastically at low FPR, and for other two models, bagging improves in some cases, but in general is not detrimental to the attack performance. 

This is particularly interesting that the attacker first doesn't need to know the training algorithm of the diffusion model, which is required for LiRA attacks, and even if the attacker chooses a small quantile regression model, which might be a weak attacker, they can still obtain nearly perfect attack performance through booststrap ensemble, namely, training a small number of small quantile regression models. 

\paragraph{Reducing Variance of Individual Predictors.} The performance improvement of bagging is mainly derived from variance reduction by taking an ensemble over several predictors. In our case, we directly take a majority vote over the decisions of individual weak attackers, and Fig. \ref{fig:bagging} already shows that overall the attack performance improves as the number of weak attackers increases. It would be interesting to show that the performance improvement indeed comes from variance reduction. 

\begin{figure*}[t]
    \centering
    \includegraphics[width=\linewidth]{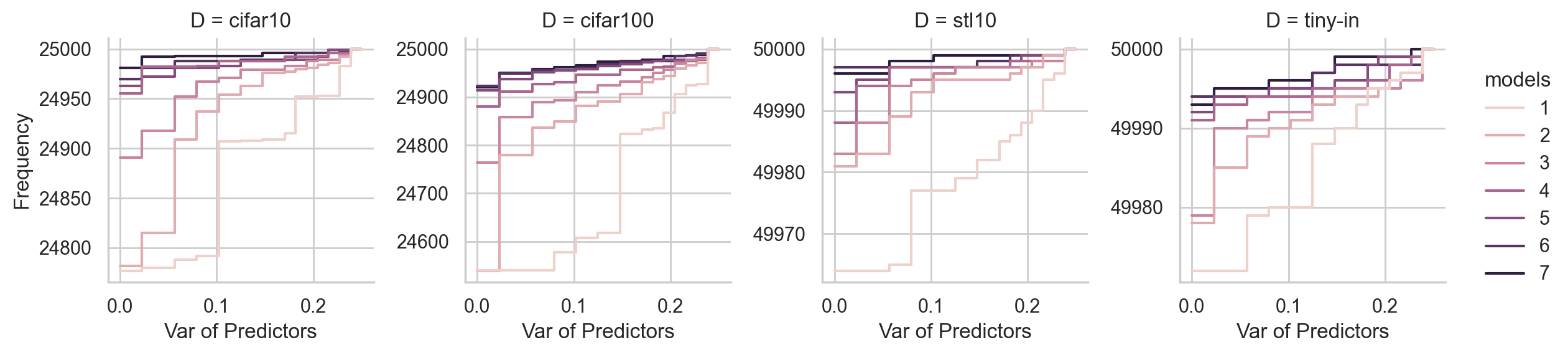}
    \caption{Variance Reduction of Bagging. We here present CDF plots of variances of bagging predictors with an increasing number of quantile regression models on four datasets in our experiments. The x-axis indicates the variance, and the y-axis indicates the cumulative count of samples used in training the diffusion model. We can see that, as the number of models increases in the bagging predictor, the CDF plot gradually becomes flatter, which means that the predictions on more samples now have lower variance. Thus, it shows that by, taking the majority vote of weak attackers,  the prediction variance decreases, which results in better performance. }
    \label{fig:variance_reduction}
\end{figure*}

We select the smallest model architecture in our experiments, which is the one with only 5,666 parameters. For a fixed number of models in bagging, we estimate the per-sample variance on the same holdout set by running our algorithm several times, and we plot the Cumulative Distribution Function (CDF) of the per-sample variance over the samples in the holdout set in 
Fig. \ref{fig:variance_reduction}. In general, as the number of models in bagging increases, the CDF shows a steeper increase at the area where variance of predictors is small. Therefore, more models in bagging lead to smaller variance.

\section{Conclusion}
We demonstrate the success of our attack based on quantile regression on diffusion models. With side information, including auxiliary samples from the data distribution, the results show the effectiveness of our membership inference attack on four diffusion models trained on different datasets respectively. Beside the performance, our attack is also computationally efficient compared to prior approaches based on shadow models. Moreover, the effectiveness and the efficiency of our algorithm indicates that diffusion models are indeed extremely vulnerable to MI attacks, and extra care should be taken when releasing a trained diffusion model. An exciting future direction is to investigate whether we can extend our current approach to a black-box attack setting, where we do not have direct access to the trained model $\theta$.

\bibliographystyle{abbrv}
\bibliography{refs.bib, Non-DP.bib}

\end{document}